\documentclass[11pt, letterpaper]{cmu}

\usepackage[all]{hypcap}
\usepackage[comma,numbers,sort,compress]{natbib}
\usepackage{hyperref}[citecolor=magenta]

\hypersetup{
    colorlinks = true,
    citecolor = {magenta},
}

\usepackage{microtype}
\usepackage{graphicx}
\usepackage{subfigure}
\usepackage{booktabs}
\usepackage{float}

\usepackage{amsmath}
\usepackage{amssymb}
\usepackage{mathtools}
\usepackage{amsthm}
\usepackage{mathrsfs}
\usepackage{nicefrac}
\usepackage{dsfont}
\usepackage{enumitem}
\usepackage{float}

\usepackage{xspace}
\usepackage[capitalize,noabbrev]{cleveref}
\usepackage{subcaption}
\usepackage{wrapfig}
\usepackage{lipsum}
\usepackage{listings}

\usepackage{amsmath}
\usepackage{amssymb}
\usepackage{mathtools}
\usepackage{amsthm}
\usepackage{bbm}
\usepackage{fleqn, tabularx}
\usepackage{multirow}
\usepackage[export]{adjustbox}
\usepackage{colortbl}

\usepackage{algpseudocode}
\usepackage{setspace}

\usepackage{color}
\definecolor{deepblue}{rgb}{0,0,0.5}
\definecolor{deepred}{rgb}{0.6,0,0}
\definecolor{deepgreen}{rgb}{0,0.5,0}
\definecolor{boost_correct_to_correct}{HTML}{66C2A5}
\definecolor{default_correct_to_correct}{HTML}{fc8d62}
\definecolor{dup_correct_to_correct}{HTML}{8da0cb}
\definecolor{new_correct_to_correct}{HTML}{e78ac3}

\newcommand\pythonstyle{\lstset{
basicstyle=\ttfamily\footnotesize,
language=Python,
morekeywords={self, clip, exp, mse_loss, uniform_sample, concatenate, logsumexp},              
keywordstyle=\color{deepblue},
emph={MyClass,__init__},          
emphstyle=\color{deepred},   
stringstyle=\color{deepgreen},
frame=single,                       
showstringspaces=false
}}

\lstnewenvironment{python}[1][]
{
\pythonstyle
\lstset{#1}
}
{}

\newcommand\pythoninline[1]{{\pythonstyle\lstinline!#1!}}

\definecolor{blanchedalmond}{rgb}{1.0, 0.92, 0.8}
\definecolor{carmine}{rgb}{0.59, 0.0, 0.09}
\definecolor{lightblue}{rgb}{0.22,0.45,0.70}

\renewcommand{\mathbf}{\boldsymbol}

\makeatletter
\def\Ddots{\mathinner{\mkern1mu\raise\p@
\vbox{\kern7\p@\hbox{.}}\mkern2mu
\raise4\p@\hbox{.}\mkern2mu\raise7\p@\hbox{.}\mkern1mu}}
\makeatother

\numberwithin{equation}{section}

\definecolor{amaranth}{rgb}{0.9, 0.17, 0.31}
\definecolor{antiquebrass}{rgb}{0.8, 0.58, 0.46}
\definecolor{antiquefuchsia}{rgb}{0.57, 0.36, 0.51}
\definecolor{chromeyellow}{rgb}{0.31, 0.47, 0.26}

\usepackage[dvipsnames]{xcolor}
\definecolor{maj5}{HTML}{2b8cbe}
\definecolor{maj5Imp}{HTML}{084081}

\definecolor{seq5wo}{HTML}{d95f0e}
\definecolor{seq5woImp}{HTML}{662506}

\definecolor{seq5w}{HTML}{6a51a3}
\definecolor{seq5wImp}{HTML}{3f007d}

\definecolor{selfwo}{HTML}{d95f0e}
\definecolor{selfwoImp}{HTML}{662506}

\definecolor{selfw}{HTML}{6a51a3}
\definecolor{selfwImp}{HTML}{3f007d}

\definecolor{glorewo}{HTML}{d95f0e}
\definecolor{glorewoImp}{HTML}{662506}

\definecolor{glorew}{HTML}{6a51a3}
\definecolor{glorewImp}{HTML}{3f007d}

\definecolor{vstar}{HTML}{d95f0e}
\definecolor{vstarImp}{HTML}{662506}

\makeatletter
\def\mathcolor#1#{\@mathcolor{#1}}
\def\@mathcolor#1#2#3{%
  \protect\leavevmode
  \begingroup
    \color#1{#2}#3%
  \endgroup
}
\makeatother

\usepackage[textsize=tiny]{todonotes}
\usepackage{algorithm}
\usepackage{amssymb}

\usepackage[skins,theorems]{tcolorbox}
\Crefformat{equation}{#2Eq.\;(#1)#3}

\Crefformat{figure}{#2Figure #1#3}
\Crefformat{assumption}{#2Assumption #1#3}
\Crefname{assumption}{Assumption}{Assumptions}

\usepackage{crossreftools}
\usepackage[suppress]{color-edits}

\usepackage{dsfont}
\usepackage{nicefrac}

\usepackage{amsmath,amssymb,amsfonts}
\usepackage{graphicx}
\usepackage{textcomp}
\usepackage{xcolor}
\usepackage{booktabs}
\usepackage{url}
\usepackage{hyperref}
\usepackage{xspace}
\usepackage{tcolorbox}
\tcbuselibrary{many}

\tcbset{
  takeawaybox/.style={
    width=\linewidth,
    top=8pt,
    bottom=4pt,
    colback=blue!6!white,
    colframe=black,
    colbacktitle=black,
    enhanced,
    center,
    attach boxed title to top left={yshift=-0.1in,xshift=0.15in},
    boxed title style={boxrule=0pt,colframe=white,},
  },
  problembox/.style={
    width=\linewidth,
    top=8pt,
    bottom=4pt,
    colback=red!6!white,
    colframe=black,
    colbacktitle=black,
    enhanced,
    center,
    attach boxed title to top left={yshift=-0.1in,xshift=0.15in},
    boxed title style={boxrule=0pt,colframe=white,},
  },
  positionbox/.style={
    width=\linewidth,
    top=8pt,
    bottom=4pt,
    colback=white!6!white,
    colframe=black,
    colbacktitle=black,
    enhanced,
    center,
    attach boxed title to top left={yshift=-0.1in,xshift=0.15in},
    boxed title style={boxrule=0pt,colframe=white,},
  },
}
\newtcolorbox{posbox}[2][]{positionbox,title=#2,#1}
\newtcolorbox{mybox}[2][]{takeawaybox,title=#2,#1}
\newtcolorbox{pbox}[2][]{problembox,title=#2,#1}

\newcounter{quotecounter}

\def\BibTeX{{\rm B\kern-.05em{\sc i\kern-.025em b}\kern-.08em
    T\kern-.1667em\lower.7ex\hbox{E}\kern-.125emX}}

\newcommand{\NN}{21\xspace}

\newcommand{\pid}[1]{%
\ifcase#1\relax
\or Client1
\or Client2
\or Resr1
\or Resr2
\or Plat1
\or Resr3
\or Plat2
\or Client3
\or Plat3
\or Resr4
\or Resr5
\or Plat4
\or Client4
\or Resr6
\or Plat5
\or Plat6
\or Resr6
\or Plat7
\or Plat8
\or Plat9
\or Plat10%
\else ERROR%
\fi
}

\newtcolorbox{analysisbox}[1][]{
    enhanced jigsaw,
    colback=white,
    colframe=blue!75!black,
    fonttitle=\bfseries,
    boxsep=5pt,
    left=5pt,
    right=5pt,
    top=5pt,
    bottom=5pt,
    title=#1,
}

\tcbset{
  aibox/.style={
    width=\linewidth,
    top=8pt,
    bottom=4pt,
    colback=blue!6!white,
    colframe=black,
    colbacktitle=black,
    enhanced,
    center,
    attach boxed title to top left={yshift=-0.1in,xshift=0.15in},
    boxed title style={boxrule=0pt,colframe=white,},
  }
}
\newtcolorbox{AIbox}[2][]{aibox,title=#2,#1}
\definecolor{lightblue}{rgb}{0.22,0.45,0.70}

\usepackage{pifont}
\usepackage{soul}
\definecolor{highlightmistake}{RGB}{255, 179, 179}
\definecolor{highlightcorrect}{RGB}{179, 255, 179}

\title{Research in Collaborative Learning Does Not Serve Cross-Silo Federated Learning in Practice}

\author[]{Kevin Kuo}
\author[]{Chhavi Yadav}
\author[]{Virginia Smith}

\affil[]{Carnegie Mellon University}

\correspondingauthor{Kevin Kuo, Chhavi Yadav (\{kkuo2,cyadav\}@andrew.cmu.edu)}

\begin{abstract}
\normalsize Cross-silo federated learning (FL) is a promising approach to enable \textit{cross-organization} collaboration in machine learning model development without directly sharing private data. Despite growing organizational interest driven by data protection regulations such as GDPR and HIPAA, the adoption of cross-silo FL remains limited in practice. In this paper, we conduct an interview study to understand the practical challenges associated with cross-silo FL adoption. With interviews spanning a diverse set of stakeholders such as user organizations, software providers, and academic researchers, we uncover various barriers, from concerns about model performance to questions of incentives and trust between participating organizations. \textit{Our study shows that cross-silo FL faces a set of challenges that have yet to be well-captured by existing research in the area and are quite distinct from other forms of federated learning such as cross-device FL.} We end with a discussion on future research directions that can help overcome these challenges.

\vspace{1cm}
\begin{tquotebox}
    {Mobile phones, that's an original use case [of FL]. But since then, I think a lot of the real world applications in academia and also industry, have been on the so-called cross-silo use case where you have different hospitals or financial institutes working together.}{\pid{9}, on real-world FL applications}
\end{tquotebox}
\begin{tquotebox}
    {While we don’t yet see many large-scale success stories of cross-silo FL, there’s growing momentum, especially in areas like the pharmaceutical industry. The main bottlenecks are often organizational or contractual rather than technical.}{\pid{16}, on industry maturity of cross-silo FL}
\end{tquotebox}
\begin{tquotebox}
    {I think people outside of [technology companies] are much more comfortable with cross-silo [FL] ... We could have separate names quite comfortably for [cross-silo and cross-device FL]. Obviously, they're very related. But the challenges are so different.}{\pid{7}, comparing cross-device vs. cross-silo FL}
\end{tquotebox}
\begin{tquotebox}
    {I feel the scope of federated learning probably could be made a bit bigger ... Because, if you think about it, nothing's model training these days.}{\pid{4}, on expanding the scope of FL}
\end{tquotebox}
\end{abstract}

\begin{document}

\maketitle
\pagebreak
\tableofcontents
\pagebreak





\section{Introduction}
Federated learning (FL) is a distributed training paradigm  that enables multiple participants to collaboratively train a model without directly sharing their data~\citep{mcmahan2017communication}. As machine learning (ML) models become an increasingly important tool and product in various industries, many organizations have turned to FL as a potential way to share data insights \textit{across organizations} while allowing each organization to retain privacy and governance of its own data. This cross-organization setting is commonly referred to as \textit{cross-silo} FL, with each organization representing a separate `data silo'~\cite{kairouz2019advances}.

\begin{figure}[b!]
    \centering
    \includegraphics[width=14cm]{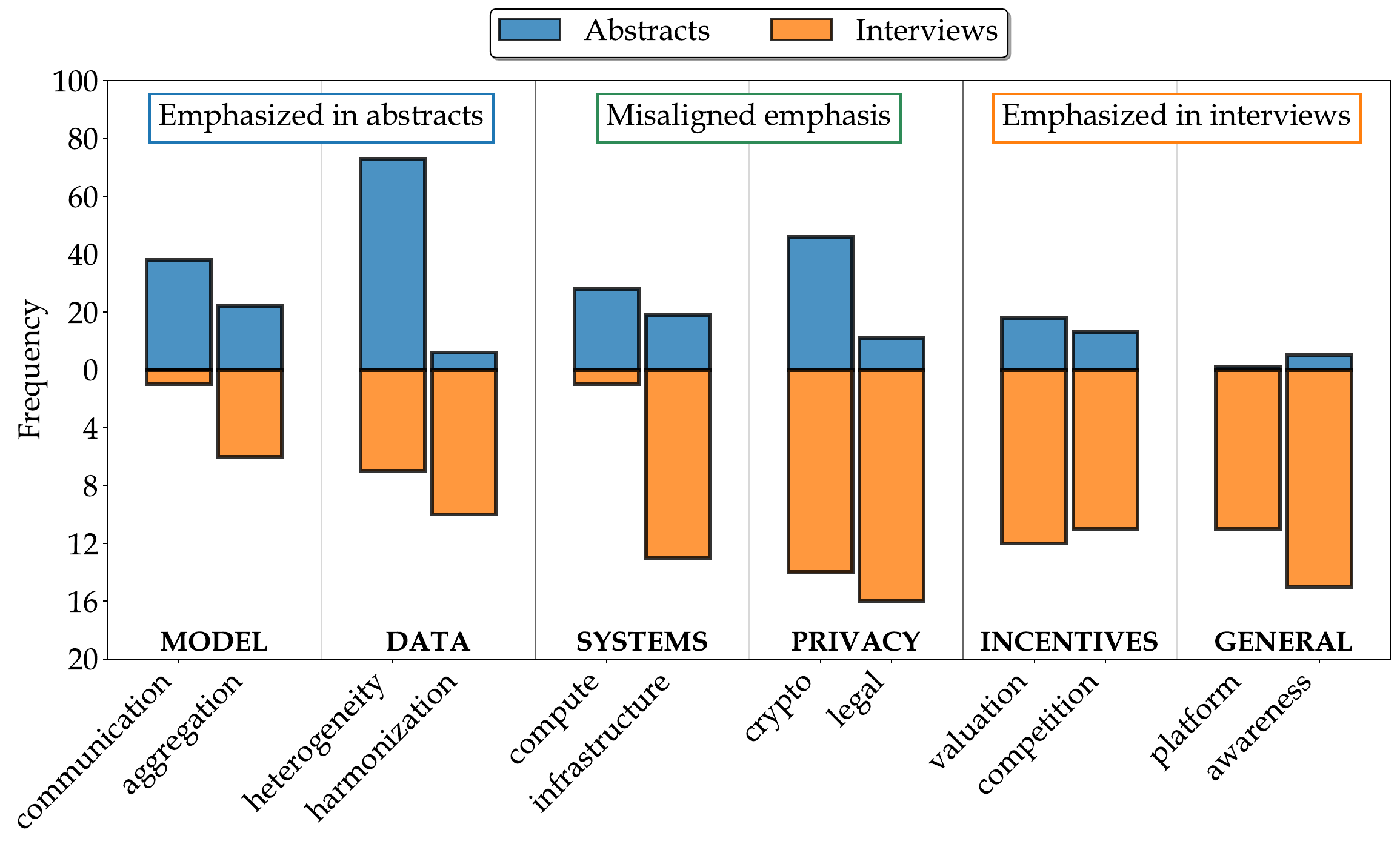}
    \caption{\textbf{Misalignment of challenges studied in Cross-silo FL research vs. those discussed in interviews}. We compare the frequency of 12 selected challenges from analyzing the top 200 cited papers on ``cross-silo federated learning" and our 21 interviews with FL stakeholders. Challenges are grouped by domain (model, data, systems, privacy, incentives, general). Literature focuses heavily on technical challenges such as model communication and data heterogeneity (left), while practitioners emphasize organizational barriers such as awareness and platform needs (right), revealing {a fundamental disconnect between academic priorities and real-world deployment challenges}. See Appendix~\ref{appendix:interview_plot} for details on how this plot was generated.}
    \label{fig:topics_abstracts}
\end{figure}

While there is significant interest in cross-silo FL,
a notable gap persists between its theoretical promise and practical deployment~\cite{huang2022cross,stricker2024fl}. We posit that a key reason for this gap is a lack of clarity surrounding the major challenges facing implementations of cross-silo FL in practice. This paper is an effort towards concretely understanding these challenges. To this end, we conduct interviews with \NN subjects who have expertise with cross-silo FL deployments, belonging to a diverse set of organizations (academia, NGOs, and private industries) and roles (researchers, engineers, managers).

The challenges identified through our interviews offer critical insights for the FL research community. In particular, our findings reveal a substantial misalignment between the challenges emphasized in cross-silo FL \textit{research} and those that are most critical in \textit{practice} (see Figure~\ref{fig:topics_abstracts}). 
For example, a significant portion of the literature on cross-silo FL focuses on developing increasingly sophisticated optimization algorithms to reduce communication overheads~\cite{pais2025strategies}, while few interview subjects mentioned communication efficiency as a major challenge. 
Conversely, there is a lack of research on evaluating legal requirements of privacy, which subjects consistently identify as a major obstacle.
These findings underscore the need for a realignment of research priorities in cross-silo FL to better address practical deployment challenges.


\vspace{.1in}
\begin{posbox}{Main position}
    {Research in cross-silo FL is 
    largely misaligned with practice, focusing on challenges that differ substantially from those that hinder real-world deployments.}
\end{posbox}
\vspace{.1in}

We hypothesize that this divergence arises from several factors. First, cross-silo FL applications are quite diverse, making it difficult to establish a clear set of research priorities for the field. Second, cross-silo FL deployments often involve organizations such as hospitals or government agencies that may lack the extensive research and marketing resources available to large technology companies. In contrast, cross-device FL has benefited from significant investment by such companies, resulting in well-defined research agendas and clearly articulated practical challenges~\cite{kairouz2019advances,bonawitz2019towards,paulik2021federated}. Third, a natural source of misalignment may stem from the implicit assumption that challenges and methodologies developed for cross-device FL directly transfer to cross-silo settings. This tendency risks overlooking the distinct characteristics of cross-silo FL and impedes its recognition as an independent area of research. 

\begin{table}[ht]

\centering
\begin{tabular}{llll}
\toprule
\textbf{Sec.} & \textbf{Challenges} & \textbf{Solutions and Open Problems} \\
\midrule
\multicolumn{3}{l}{\textbf{Organizational}} \\
\midrule
\S\ref{sec:ideation} & (\ref{quote:ideation-awareness},\ref{quote:data-maturity}) Lack of awareness/maturity & (\ref{quote:ideation-formation},\ref{quote:awareness}) Platform responsibilities/outreach \\
& (\ref{quote:ideation-systems}) Inertia against switching systems & (\ref{quote:ideation-approach}) Co-design initiatives \\
& (\ref{quote:ideation-fail}) Establishing a business use case & (\ref{quote:ideation-needs}) Guidelines to assess FL feasibility \\
\S\ref{sec:prototyping} & (\ref{quote:feature-names}) Building a common data pipeline & Domain-specific data standards and tools \\
\S\ref{sec:evaluation} & (\ref{quote:data-volume}) Fair/equal client contributions & (\ref{quote:incentives-valuation}) Require similar training data quantity\\
& (\ref{quote:eval-centralized}) Unclear evaluation standards & Auditing tools; (\ref{quote:eval-dp}) comprehensive tests \\
& (\ref{quote:eval-communication}) Meeting privacy regulations & (\ref{quote:regulator}) Fragmented privacy laws \\
& & Cross-domain ML and privacy law research \\
& (\ref{quote:trust-star}) Establishing trust among parties & (\ref{quote:trust-circle}) Central coordinator; (\ref{quote:eval-system}) Legal contracts \\
& (\ref{quote:incentives-outcomes}) Convincing clients to participate & (\ref{quote:eval-interface}) Behavioral and HCI research in FL \\
& & (\ref{quote:scope}) Expanding the scope of FL \\
& (\ref{quote:trust-scale}) Cost of establishing standards & Network growth; involving smaller clients \\
\midrule
\multicolumn{3}{l}{\textbf{Technical}} \\
\midrule
\S\ref{sec:prototyping} & (\ref{quote:data-heterogeneity}) Heterogeneity in production data & Identify/predict impact of heterogeneity \\
& (\ref{quote:data-guidance}) Limited data for prototyping & Leveraging public and synthetic data \\
& (\ref{quote:poc-ease}) Ease of system prototyping & Development libraries; debugging tools \\ 
\S\ref{sec:evaluation} 
& (\ref{quote:trust-blockchain}) FL with untrusted parties & Cryptographic methods \\ 
\S\ref{sec:deployment} & (\ref{quote:deploy-expertise}) Setup needs diverse expertise & (\ref{quote:deploy-access}) Automated configuration tools \\
& (\ref{quote:deploy-monitor}) System monitoring and updating & MLOps for FL; (\ref{quote:modular}) Opt-out guarantees \\
\bottomrule
\end{tabular}
\caption{Challenges, Solutions and Open Problems}\label{tab:challenges}
\end{table}

The rest of the paper is organized as follows. In Section~\ref{sec:interview}, we discuss the key stakeholders in the cross-silo FL ecosystem, our participant recruitment process, and our interview protocol. In Section~\ref{sec:findings}, we detail the main findings, organized around the different stages of the FL pipeline. Table~\ref{tab:challenges} contains a comprehensive overview of the challenges raised in our interviews and their associated quotes. Lastly, in Section~\ref{sec:discussion}, we conclude with a discussion on open problems and potential solutions, along with a reflection on the underlying causes of the persistent gap between research priorities and practical barriers in cross-silo FL.

\label{sec:intro}

\section{Background \& Related Work}
\label{sec:relwork}
\begin{figure}[b!]
    \vspace{-.2in}
    \centering
    \includegraphics[width=14cm]{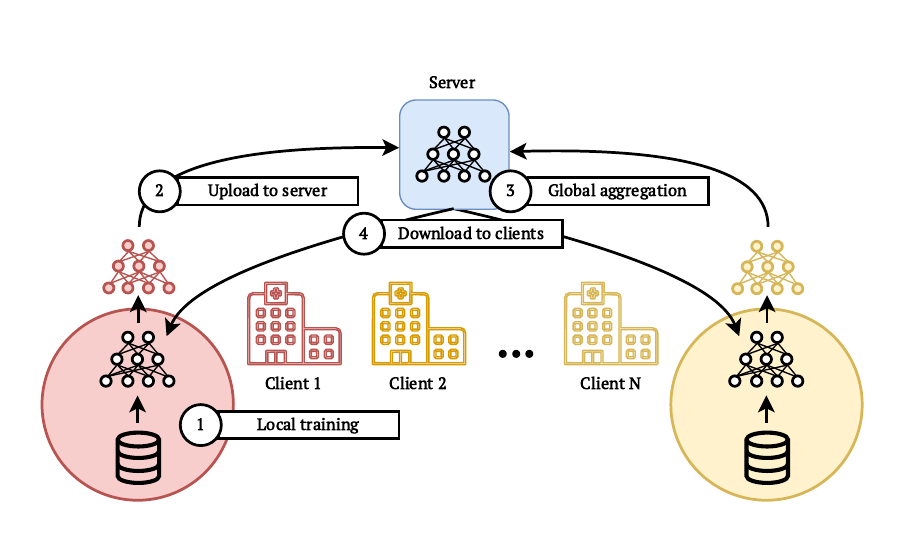}
    \vspace{-.1in}
    \caption{\textbf{An example diagram of a cross-silo FL training procedure.} Each client is an \textbf{organization} with its own \textbf{data silo}. At each communication round, (1) clients perform local training, and (2) upload the trained model to the server. The server then (3) aggregates the client models into a global model. Lastly, clients (4) download the new aggregate model and repeat this process.}
    \label{fig:fl_algorithm}
\end{figure}

\textbf{Federated and collaborative learning.} \textit{Federated learning}~\cite{mcmahan2017communication} involves training a machine learning model in a distributed fashion---most commonly using a central server to communicate with clients (each storing their own local dataset) in a network with a star topology, as shown in Figure~\ref{fig:fl_algorithm}. Training is often performed in several synchronous training rounds; that is, all clients download the global model, perform a predetermined amount of local training, and then upload their model updates to the server. The next round only starts after the server has received and aggregated all the updates. 
There exist multiple variants of FL which modify this setting in various ways, such as asynchronous~\cite{wu2020safa,liu2021adaptive,wang2022asynchronous,nguyen2022federated,xu2023asynchronous,yu2023async}, hierarchical~\cite{liu2020client,abad2020hierarchical,liu2022hierarchical,ooi2023measurement}, and decentralized FL~\cite{qu2022decentralized,gabrielli2023survey,beltran2023decentralized,yuan2024decentralized}. 
Areas of FL research include the exploration of issues such as data heterogeneity and personalization
\cite{li2020federated,gao2022end,wu2022motley,cheng2023protohar,huang2023generalizable,miao2023fedseg,kang2022fedcvt,kang2022privacy,zhou2024personalized,cho2023communication,kang2022communication,bao2023optimizing,qi2024fdlora,chen2023fedsoup,zhou2023hierarchical,feng2024robustly}, cross-domain and transfer learning
\cite{wang2021federated,qian2024heterogeneous,parekh2021cross,wang2021poi,yang2022cross,yan2024cross,yang2024cross,yu2023multimodal,majeed2021cross,das2022cross,liu2024vertical}, privacy and security
\cite{wen2022fishing,suri2022subject,tian2024privacy,chamikara2022local,hahn2021versa,yang2022practical,gong2023multi,liu2021revfrf,lowy2023private,lowy2023private2,wang2024can}, compression
\cite{gong2022preserving,han2022fedx,nguyen2022novel,wang2024aggregation,li2024filling}, communication efficiency
\cite{jin2021cross,zhou2023every,wang2021efficient,ro2022scaling,dorfman2023docofl,li2023convergence}, systems
\cite{chen2023fs,gao2023federated,li2020esync,stripelis2022semi,zhang2023timelyfl,liu2022multi,patros2022rural,wang2023flint,kuo2023noisy,stricker2024fl}, blockchain
\cite{ur2021trustfed,kang2023blockchain,rahmadika2021blockchain,majeed2021st,ranathunga2022blockchain}, 
and other specialized methods and applications
\cite{li2022federated,jianping2024federated,rasti2022graph,rehman2023dawa,xu2024federated,bey2020fold,vzalik2023review,zhu2022cross,wang2022fed}. There are also other settings that may be considered under the umbrella of collaborative learning, such as distributed learning or split learning~\cite{gupta2018distributed,vepakomma2018split}. Our interview findings extend to these setups in a cross-organizational context, due to similar challenges around systems, privacy, and incentives. 




\textbf{Cross-Device and Cross-Silo FL.} FL settings can be loosely categorized as either \textit{cross-device} or \textit{cross-silo}~\cite{kairouz2019advances}. The most prominent examples of production FL systems are in the \textit{cross-device} setting, where each user is associated with a person or device and the network contains a large number of such clients. Such systems are deployed by technology companies who have sole access to large-scale user networks~\citep{yang2018applied,bonawitz2019towards,paulik2021federated,stojkovic2022applied,huba2022papaya,ji2025private}. Meanwhile, \textit{cross-silo} FL is more generally applicable to small-to-medium sized networks where each client represents an organization or data silo. This setting is the focus of our study.


While cross-silo FL has received less attention relative to cross-device FL, some studies have brought key challenges to light, such as incentives~\cite{tang2021incentive,karimireddy2022mechanisms,cho2022federate,kong2022incentivizing,zhang2022enabling,zeng2022incentive}, privacy \cite{heikkila2020differentially,li2021practical,lomurno2022sgde,liu2022privacy}, encryption \cite{jiang2021flashe,luo2022svfl,tran2023efficient,ma2025armadillo}, 
and real-world deployment~\cite{dayan2021federated,heyndrickx2023melloddy,soltan2024scalable,asti2024artificial}. {We explore these previously identified concerns and found that while they indeed overlap with those in practice, stakeholders often focused on factors which are underexplored in the literature, as illustrated in Figure~\ref{fig:topics_abstracts}.

\textbf{FL Surveys.} Several literature surveys have been conducted on general~\cite{kairouz2019advances,li2020survey,wen2023survey,liu2024vertical,pei2024review,woisetschlager2024survey,daly2024federated}, domain-specific~\cite{rahman2023federated,joshi2022federated,teo2024federated,eden2025scoping,guan2024federated,shi2023responsible,vzalik2023review}, and cross-silo FL settings~\cite{huang2022cross,huang2023promoting,korneev2025survey}. In contrast to these surveys, we conduct an interview study aimed at understanding challenges of real-world cross-silo FL systems. Our study shares a practical focus with recent work discussing the priorities of FL under data regulations such as the EU AI Act~\cite{woisetschlager2024priorities,woisetschlager2024responsible}. One key contribution of our work is comparing challenges raised in interviews to those which are studied in the literature, helping to guide future research directions to better serve cross-silo FL in practice.

\textbf{ML Interview Studies.} Other studies have interviewed ML researchers and non-ML domain experts on topics such as adoption of ML systems in clinics~\cite{pumplun2021adoption}, perspectives on ethical issues of clinical ML tools~\citep{kim2023physicians}, and operationalizing machine learning~\cite{shankar2024we}. Similar to these studies, we seek to understand ML systems from a practitioner's perspective. However, to the best of our knowledge, no prior study has attempted to interview cross-silo FL practitioners. Furthermore, our study focuses on the diversity of stakeholders and their contributions throughout different stages of the FL pipeline.

\section{Interview Methods}
\label{sec:interview}
\textbf{Subject Recruitment.} 
During recruitment, we searched for contacts representing three types of \textbf{stakeholders} categorized by their \textbf{organization's} role in the FL ecosystem.

\vspace{-.1in}
\begin{itemize}
    \item \textbf{Client stakeholders} are from (a consortium of) organizations which work on an end application, often in healthcare or finance. These organizations are \textbf{data owners} and have downstream applications for this data. Ultimately, FL adoption depends on the willingness of \textbf{decision-makers} at these client organizations.
    \item \textbf{Platform developers} provide infrastructure to facilitate FL and other collaborative AI initiatives between multiple client organizations. Infrastructure services can vary greatly across platforms; some only provide software libraries, while others provide on-site hardware support.
    \item \textbf{Researchers} provide technical assistance with ML and FL aspects of the project. Their organization may have links to industry, but does not directly work with client organizations. For example, this may be a platform in early development or a research institution collaborating with a platform.
\end{itemize}
\vspace{-.1in}

We recruited subjects with a broad set of \textbf{backgrounds}, such as medical professionals, system engineers, product managers, and FL researchers. 
These subjects work at a variety of organizations, such as hospitals, banks, academic research labs, and software startups. We categorize subjects by the \textbf{application} their organization is focused on:

\vspace{-.1in}
\begin{itemize}
    \item \textbf{ML Research}: refers to fundamental algorithms for distributed and collaborative learning, and does not (yet) target a particular domain or business application.
    \item \textbf{Healthcare}: includes applications such as drug discovery, medical imaging, and measuring population outcomes.
    \item \textbf{Finance}: includes fraud and anomaly detection applications.
    \item \textbf{Various}: refers to platforms targeting multiple domains.
\end{itemize}
\vspace{-.1in}

A description of the \NN total subjects is shown in Table~\ref{tab:subjects}. We recruited and interviewed subjects on a rolling basis over a 5-month period (May to September 2025). 

\textbf{Interview Protocol.}
We conducted semi-structured interviews over Zoom, each lasting 30-60 minutes. Each subject received a written consent form before the interview and agreed to participate free of compensation. For users who consented to being recorded, we transcribed and recorded the the interview audio using Zoom's built-in features. This was an IRB-approved study. 

During the interview, we asked a series of open-ended questions that broadly spanned (1) the subject's role at their organization, (2) how their organization uses or seeks to use cross-silo FL, (3) the process and current state of cross-silo FL adoption in their organization, (4) any barriers in deploying an end-to-end cross-silo FL system. Due to the diverse organizations and roles of the interview subjects, we tailored the questions based on their role. A full list of guiding questions is provided in Appendix~\ref{appendix:questions}.

\begin{table*}[h]

\centering
\begin{tabular}{lllllll}
\toprule
\textbf{ID} & \textbf{Background} & \textbf{Organization} & \textbf{Application} & \textbf{Stage} & \textbf{Site} & \textbf{Quotes} \\
\midrule
\pid{1} & Statistics, Public Health & Client & Healthcare & Prod & US-East & \\
\pid{2} & Oncology & Client & Healthcare & Prod & US-West & \\
\pid{3} & LLMs & Researcher & ML Research & POC & US-West &\ref{quote:modular}  \\
\pid{4} & LLMs, Edge AI & Researcher & ML Research & POC & US-East & \ref{quote:eval-interface}, \ref{quote:scope}  \\
\pid{5} & Management & Platform & Healthcare & Prod & US-East &\ref{quote:ideation-awareness}, \ref{quote:awareness}  \\
\pid{6} & Efficient ML & Researcher & ML Research & POC & US-East & \ref{quote:ideation-systems}  \\
\pid{7} & FL & Platform & Various & Prod & Intl &\ref{quote:poc-ease}, \ref{quote:eval-centralized}, \ref{quote:naming}  \\
\pid{8} & Cryptography & Client & Finance & Prod & US-East &\ref{quote:feature-names}, \ref{quote:trust-star}  \\
\pid{9} & FL & Platform & Various & Prod & US-East &\ref{quote:ideation-needs}, \ref{quote:applications}  \\
\pid{10} & FL & Researcher & Healthcare & Prod & Intl & \\
\pid{11} & FL & Researcher & Finance & POC & Intl &\ref{quote:ideation-fail}, \ref{quote:data-guidance}, \ref{quote:resources}  \\
\pid{12} & AI, Data Science & Platform & Healthcare & POC & Intl &\ref{quote:deploy-access}  \\
\pid{13} & AI, Data Science & Client & Finance & POC & US-East &\ref{quote:data-heterogeneity}, \ref{quote:trust-blockchain}, \ref{quote:deploy-monitor}  \\
\pid{14} & Decentralized ML & Researcher & ML Research & POC & Intl & \\
\pid{15} & Management & Platform & Various & Prod & Intl &\ref{quote:ideation-approach}, \ref{quote:deploy-expertise}  \\
\pid{16} & Biochemistry & Platform & Healthcare & Prod & Intl &\ref{quote:ideation-formation}, \ref{quote:data-maturity}, \ref{quote:incentives-valuation}  \\
\pid{17} & ML, Privacy & Researcher & Various & POC & US-East &\ref{quote:eval-dp}, \ref{quote:deploy-compute}  \\
\pid{18} & Decentralized ML & Platform & Various & Prod & Intl & \\
\pid{19} & Management & Platform & Healthcare & POC & US-West & \\
\multirow{2}{*}{\pid{20}} & \multirow{2}{*}{Statistics, Genomics} & \multirow{2}{*}{Platform} & \multirow{2}{*}{Healthcare} & \multirow{2}{*}{Prod} & \multirow{2}{*}{Intl} & \ref{quote:data-volume}, \ref{quote:eval-communication}, \ref{quote:trust-circle} \\
& &  &  &  &  & \ref{quote:trust-scale}, \ref{quote:regulator} \\
\pid{21} & AI, Data Science & Platform & Healthcare & POC & Intl &\ref{quote:eval-system}, \ref{quote:incentives-outcomes}  \\
\bottomrule
\end{tabular}
\caption{Organization Roles and Applications}\label{tab:subjects}
\end{table*}

\textbf{Subject Naming.}
Here, we clarify the meaning and usage of the naming terms we use in Section~\ref{sec:findings}. In general, we use the term ``subjects'' as shorthand for ``interview subjects''. We avoid using the term ``participant'' due to the ambiguity between ``interview participant'' vs ``FL network participant''. We use ``stakeholder'' to broadly refer to any individual involved in an industry cross-silo FL project. For the terms ``client'', ``platform'', and ``researcher'', these are used to capture both the organizational \textbf{affiliation} of stakeholders and their degree of \textbf{involvement} in FL applications. Therefore, it is important to note that these terms do \textbf{not} necessarily capture the stakeholder's organizational role or expertise. Terms such as ``engineer'', ``developer'', and ``practitioner'' are used to describe stakeholders with technical expertise in computer systems, ML, or FL, and may be from any of the three types of organizations.
Finally, in some cases, it is ambiguous whether these terms refer to an individual or organization. We use ``client'' and ``platform'' to refer to organizations, while ``researcher'' refers to individuals but can be interpreted as either. Otherwise, we specify additional terms e.g. ``client stakeholder'' or ``client organization''.
\section{Cross-Silo FL Challenges In Practice: Our Findings}
\label{sec:findings}

In this section, we present the findings from our interviews, organized according to the stages of the cross-silo FL pipeline for clarity of exposition. Specifically, we structure our analysis around the following stages: \textbf{Ideation} (\ref{sec:ideation}) is an initial stage where clients develop a general interest in FL and assess its value for their application.
\textbf{Prototyping} (\ref{sec:prototyping}) involves system design, method implementation, and FL simulation.
\textbf{Evaluation} (\ref{sec:evaluation}) is the process of analyzing prototypes and potentially convincing decision makers.
\textbf{Deployment} (\ref{sec:deployment}) is the final stage where the FL system is set up and run at remote sites.

At a high level, our findings share three common themes across interviews, which correspond to the discrepancies between cross-device and cross-silo FL identified in Section~\ref{sec:intro}:

\textit{\textbf{Heterogeneity creates multiple early challenges.}} In~cross-silo FL, the most common failure points involve negotiation between clients on issues like infrastructure cost, privacy, and incentives in a way which satisfies each client's heterogeneous requirements. Heterogeneity is not only an issue for client requirements within the same network, but also for network requirements across different applications. This is because cross-silo FL applications vary in how to address the intra-network issues mentioned above; therefore, the conditions for cross-silo network formation are not easily repeatable. 
We note that these forms of heterogeneity are distinct from concerns around `data heterogeneity' in the FL research literature, which consider differing data distributions at each silo (Figure~\ref{fig:topics_abstracts}).

\textit{\textbf{Underinvestment raises barriers to entry.}} Organizations which take an interest in cross-silo FL tend to have limited resources compared to the large companies deploying cross-device FL. Therefore, cross-silo FL challenges   the project at any stage. Ideation requires baseline awareness about cross-silo FL and data-driven practices. Prototypes need minimal client buy-in, yet must demonstrate value. Evaluation is complex and lacks standards. Lastly, the engineering costs of cross-silo FL systems are expensive. 
Due to these barriers, organizations need strong reasons to adopt cross-silo FL.

\textit{\textbf{A lack of openness impedes progress.}} For emerging technologies like FL to succeed, it is critical for research objectives to match business applications. Unfortunately, a lack of relevant and open information between industry and research makes this difficult. Subjects noted that practitioners lack clear guidance on how research methods apply to their contexts, while academics lack visibility into real-world deployment challenges.

\textbf{For brevity, in the rest of the paper we use ``FL'' to mean ``cross-silo FL'' unless specified otherwise.}

\subsection{Ideation}
\label{sec:ideation}
The first stage of a (cross-silo) FL pipeline is \textbf{ideation}, where a client (organization) must assess the value of an FL system. When clients have a strong interest in FL, decision-makers in these companies can drive projects in a top-down manner, simplifying future challenges. Next, we identify several underlying challenges that affect the willingness of organizations to pursue FL projects.  

\textbf{Clients generally lack awareness of FL.} A fundamental barrier to FL adoption is lack of general awareness. The definition of FL can be unclear even among ML researchers. The word `federated' is often conflated with related terms such as `distributed', `decentralized' or `edge', due to overlapping techniques and applications. Furthermore, clients are less familiar with `federated' compared to similar terms (\ref{quote:ideation-awareness}). 
Additionally, subjects from various organizations brought up challenges of communicating technical concepts to client stakeholders such as medical and IT professionals. These areas not only included FL, but also privacy, secure computation, and model evaluation (\pid{2}, \pid{4}, \pid{5}, \pid{8}, \pid{11}).

\refstepcounter{quotecounter}\label{quote:ideation-awareness}
\begin{quotebox}
    {There's a ton of education that has to happen [around FL]...if you say edge computing, people know what that means, which is interesting ... the `federated' word can trip people up.}{\pid{5}, on client awareness of federated learning}
\end{quotebox}

In other cases, organizations who are eager to set up FL systems generally have a strong understanding of these systems and have realistic expectations for both their risks and added value. Here, subjects reportedly had little difficulty convincing their clients to participate in FL, as the clients had already decided they wanted to use FL (\pid{10}) or were compensated for participating in the research program (\pid{12}). 


\textbf{Costs and awareness affect client inertia.} 
Even when FL is a promising approach and is not hindered by regulations, decision-makers may still be unwilling to pursue FL solutions due to the costs needed to migrate to a new system. Similar to ML systems, FL systems require a large amount of engineering labor to set up the data pipeline and compute infrastructure. However, in the context of FL, there are additional challenges with these tasks due to different system requirements across client organizations. In \ref{quote:ideation-systems}, \pid{6} discussed how non-traditional collaborative systems are difficult to set up, and that users ultimately prefer to use their own systems with existing investment. 

\refstepcounter{quotecounter}\label{quote:ideation-systems}
\begin{quotebox}
    {...people actually need to be willing to participate ... people ask themselves, ``Okay, and I can use all these other things that already work perfectly. Why should I use your system?''}
    {\pid{6}, on the challenge of client inertia}
\end{quotebox}



\textbf{Success starts with aligned interests.}
In \ref{quote:ideation-fail}, \pid{11} suspected that the main reason for the failure of their project was due to a lack of interest from the business side. While clients have enough interest to develop a proof-of-concept, they often lack the motivation to bring the project to deployment. 

\refstepcounter{quotecounter}\label{quote:ideation-fail}
\begin{quotebox}
    {It would have been interesting, and also not that hard to get it done. But I think the interest in actually delivering a product in federated learning came at some point more from us as researchers than from the company. And that's not the way it should be, and I think it led to the project ultimately failing.}
    {\pid{11}, on how FL projects need client interest}
\end{quotebox}

Overall, FL projects which are driven by an immediate and practical need for data tend to succeed the most, compared to projects which are initiated from the context of research. In \ref{quote:ideation-needs}, \pid{9} attributed their success to the goals of their project matching concurrent needs in healthcare.

\refstepcounter{quotecounter}\label{quote:ideation-needs}
\begin{quotebox}
    {So in healthcare, I think we saw a need. And we kind of built this technology to solve that need and provide a solution. I think, for other use cases, sometimes they are just not aware of this type of technology being available.}
    {\pid{9}, on why their project was able to succeed}
\end{quotebox}

Therefore, it is important for decision-makers to be knowledgeable about FL systems and convince their organizations that setting up such a system is feasible and worth the cost. \pid{16} also noted that the maturity of an organization's data-driven processes is a prerequisite for FL awareness. During the COVID-19 pandemic, for example, many healthcare organizations improved their practices around data-driven processes (\pid{9}, \pid{20}); as a result, this particular combination of regulatory constraints and application needs has pushed the healthcare industry to pursue FL among other privacy-enhancing technologies.


\textbf{Platforms can help keen but inexperienced clients.} FL adoption ultimately depends on the willingness of decision-makers who often lack technical expertise (\pid{11}, \pid{18}). To address challenges with awareness, several platforms (\pid{5}, \pid{7}, \pid{15}) pointed to active collaborative efforts, such as on the project-level by having external engineers visit the client site, or as part of a broader outreach program. In \ref{quote:ideation-approach}, \pid{15} discussed how platform-client collaborations form; clients must usually have a base level of interest around FL before they approach a platform, while preemptively approaching clients is unproductive. To lower this barrier to entry, \pid{15} also discussed a co-design program intended for clients with little to no prior experience with FL.

\refstepcounter{quotecounter}\label{quote:ideation-approach}
\begin{quotebox}
    {It doesn't really serve to approach a hospital directly about FL, unless they already have, you know, a federated learning stream of work underway, [or] medical professionals in there who are also researchers and who are already keen on FL. We do get quite a bit of inbound ... that's what the co-design program is for.}{\pid{15}, on a co-design program to help lower the organizational barrier to FL}
\end{quotebox}

Platforms can also play a critical role in establishing coalitions. In particular, \pid{16} discussed how their platform 
not only provides technical support, but also takes on important central responsibilities throughout the collaboration, such as contracting and deployment~(\ref{quote:ideation-formation}).

\refstepcounter{quotecounter}
\label{quote:ideation-formation}
\begin{quotebox}
    {So [in several networks], we're really the coordinator. We take care of contracting, deployment, all the organizational things that come about from coordinating multiple partners, so there we have really central roles.}
    {\pid{16}, on the many coordination tasks their platform performs for clients}
\end{quotebox}


\begin{mybox}{Takeaway}
\textbf{Ideation} of FL solutions is driven by external regulatory constraints and application requirements, the readiness to adapt a new technology, as well as the internal preferences of innovators and decision-makers within an organization.
\end{mybox}
\subsection{Prototyping}
\label{sec:prototyping}
In order to convince decision-makers, stakeholders often need to develop a proof-of-concept that demonstrates the value of FL. Subjects reported a variety of perspectives on this \textbf{prototyping} stage, depending on the client's readiness to use FL. In less mature cases, prototyping simply helps to familiarize the decision-makers and engineers at the client organization with FL and assess whether it is a suitable solution. In more mature cases, organizations spend this time developing software to meet specific expectations (for instance the FL algorithm, model architecture, and data format) depending on the application and client infrastructure. Next we discuss our findings from this stage.

\textbf{FL projects need a distributed data pipeline.}
To develop cross-silo FL applications, it is necessary to have a common data architecture before any modeling can begin. When asked about initial development challenges, subjects (\pid{1}, \pid{7}, \pid{8}, \pid{13}, \pid{16}) started with basic concerns, such as what format files are stored in or what names are used for various features (\ref{quote:feature-names}). While these concerns are simple from a technical view, they can be challenging from a logistical and coordination view as they may require an additional layer of approval for clients to release their data model or a sample of their data.

\refstepcounter{quotecounter}\label{quote:feature-names}
\begin{quotebox}
    {Instead of `address', [another client] will call it `location'. I mean, I'm just giving you an example, you need to do this exercise at the beginning to find out, like how to make the data consistent with each other. And I think that's easier internally, I mean, externally, we've not done it, but I'm imagining that it will be much harder.}
    {\pid{8}, on different names for data features}
\end{quotebox}

Subjects (\pid{7}, \pid{9}, \pid{10}) reported that data harmonization was a commonly requested feature. Current harmonization tools validate and transform data into standardized formats for public or private repositories. For example, harmonization tools for genomics applications focus on transforming ``contextual data'' such as ``laboratory, epidemiological, and methodological information'' into formats that comply with ARS-CoV2 data standards~\citep{gill2023dataharmonizer}.
Lastly, in settings where organizations have different methods or environments for data collection, subjects mentioned that it was important to perform data normalization in a privacy-preserving way (\pid{10}, \pid{16}). 

\begin{pbox}{Open Problem}
    {{FL research should consider challenges of system components outside of model training, as these components (e.g. data pipelines) can face early issues of coordination and privacy.}}
\end{pbox}

\textbf{Data heterogeneity is important but not critical.} Subjects were aware that data heterogeneity can arise in various applications and that it may also impede FL model training. For example, \pid{2} discussed that heterogeneity is present at multiple scales when performing CT scans across different hospitals, ranging from obvious (the number of scans) to subtle (how long technicians wait between contrast dye injections). \pid{3} and \pid{6} also brought up challenges with out-of-domain generalization and optimizing models across heterogeneous data sources. For example, mixture-of-experts models are a promising method to effectively combine expert knowledge from distinct domains, but degrade as the number and specificity of experts increases beyond a certain point~\cite{li2022branch,sukhbaatar2024branch,shi2025flexolmo}. 

While there was general awareness and concern about data heterogeneity, client perspectives were varied. Some did not mention data heterogeneity at all, while some subjects did not mention its impact on FL training and instead brought up more general aspects. For example, \pid{1}, and \pid{19} noted that certain populations (e.g. associated with a unique region or disease) can be more difficult to model and having more diverse data can address this issue. Other subjects (\pid{13}, \pid{17}) who had more hands-on experience with FL stated that it is important to simulate data heterogeneity in prototypes using public or synthetic data (\ref{quote:data-heterogeneity}). 

\refstepcounter{quotecounter}\label{quote:data-heterogeneity}
\begin{quotebox}
    {We tried to sample the distributions to reflect the fact that the banks would be located on separate continents.}
    {\pid{13}, on modeling data heterogeneity in a financial application of FL}
\end{quotebox}


However, in a similar number of discussions, subjects focused on data quantity as the main source of heterogeneity~(\ref{quote:data-volume}). Furthermore, these stakeholders tended to have an incentives-based perspective on this imbalance in data, i.e., that clients with less data unfairly benefit more from FL (\S\ref{sec:evaluation}). This is in contrast to academic literature on FL, which most often studies more general/complex forms of data heterogeneity, and considers the effect on the overall quality of the final model.

\refstepcounter{quotecounter}\label{quote:data-volume}
\begin{quotebox}
    {[Concerns of data heterogeneity] did come up, but not so much about, like, which data, but the volume of data.}
    {\pid{20}, on user concerns of data imbalance}
\end{quotebox}

Interestingly, very few subjects were explicitly concerned about the effect of data heterogeneity on model quality. Even in these cases, they did not have clear solutions beyond simply including more diverse sources of data or tuning training hyperparameters and rarely mentioned algorithmic solutions. Overall, this indicates that even though academic FL research proposes many solutions for data heterogeneity, solutions that have made an impact in practice consider limited definitions of heterogeneity (e.g. data quantity and label frequency). Another interesting disconnect is that these papers begin by presenting data heterogeneity as the central issue for poor model performance. However, outside of simulation settings, it is difficult to attribute poor model performance to data heterogeneity and quantify how problematic data heterogeneity is. Therefore, there is a need for more general papers on identifying data heterogeneity and studying how it  occurs naturally~\cite{wu2022motley}.

\begin{pbox}{Open Problem}
\label{problem:data_heterogeneity}
    {While FL research focuses heavily on addressing data heterogeneity, there is a lack of research on identifying/characterizing heterogeneity in practice and predicting its impact.}
\end{pbox}

\textbf{Public and synthetic data can address data scarcity.} Data is scarce for many FL applications, making it difficult to create meaningful prototypes. A common theme across interviews was that organizations interested in FL have small datasets and expect significant benefits from having access to more data. 
Beyond data quantity, subjects also specifically mentioned the need for data from more diverse populations (\pid{2}, \pid{3}, \pid{18}, \pid{19}) or to link records to additional features (\pid{20}).

Public and synthetic data play an important role in developing proof-of-concepts in both research and practice (\pid{3}, \pid{7}, \pid{8}, \pid{11}). As high-quality data is scarce in cross-silo FL applications, developers rely on data owners to provide synthetic data or tools to generate it (\pid{13}, \pid{15}, \pid{17}). In several cases, proof-of-concepts only used public data, due to the early stage of the project and a lack of cooperation from data owners (\ref{quote:data-guidance}). Furthermore, subjects (\pid{11}, \pid{13}, \pid{17}, \pid{20}) noted that even with a prototyping dataset, simulating data heterogeneity in a meaningful way can be difficult. Due to limited knowledge about how heterogeneity might occur in their data application, subjects often simulated heterogeneity only in terms of data quantity and label distribution. In the best case scenarios, those developing the proof-of-concept either had the domain expertise to simulate domain-specific forms of data heterogeneity or worked with data providers who were willing to generate synthetic data. In some cases, due to restrictions on data, developers must test proof-of-concept systems on an entirely different domain than what is expected in production.

\refstepcounter{quotecounter}\label{quote:data-guidance}
\begin{quotebox}
    {We just looked at common [fraud rate] industry numbers from interviews, because we wouldn't even get these from the company we're working with.}
    {\pid{11}, on a fraud detection proof-of-concept}
\end{quotebox}


Due to the challenges posed by data scarcity, there is a need for more research on both curating public FL datasets~\cite{ogier2022flamby} and how to leverage public data for FL. For example, as discussed in Section~\ref{sec:relwork}, many existing works study public data in the context of pretraining or knowledge distillation. Future studies should build on these works to address challenges of communication, computation, and privacy~\cite{kuo2023noisy,hou2024pre,wang2024can}.

\begin{pbox}{Open Problem}
\label{problem:data_public}
    {To tackle data-scarce FL applications, it is important to develop methods that leverage public data in the context of FL and consider broader analyses of when public data is expected to help.}
\end{pbox}

\textbf{Researchers and platforms support tools and standards.} A critical role of platforms is to offer software tools to assist with prototyping. This serves multiple essential functions: reducing engineering costs, demonstrating value to stakeholders, and validating system functionality. Subjects (\pid{7}, \pid{8}, \pid{11}) emphasized the importance of lowering the barrier to entry by offering compatible and easy-to-use tools, such that a single ``champion'' can easily present their results~(\ref{quote:poc-ease}). On the technical side, several subjects also noted that debugging distributed systems can be difficult, and therefore having tools to validate their own system was useful. For example, \pid{7} discussed tools to catch and debug silent FL errors, while \pid{15} and \pid{20} discussed ``sandbox'' environments. These sandboxing tools allow practitioners to interact with mock data or models in an isolated environment to both validate the functionality of code as well as perform tests on sensitive data.

\refstepcounter{quotecounter}\label{quote:poc-ease}
\begin{quotebox}
    {...this is a disruptive, emerging type of technology. So you have to build it in a way that a single champion can demo it on their laptop, show their boss, or convince an important stakeholder...We basically think if we give people enough tools, the stuff will start to take off because there's enough interest.}
    {\pid{7}, on the importance of easy-to-use FL tools}
\end{quotebox}

\begin{mybox}{Takeaway}
While \textbf{prototypes} help demonstrate the value of FL to decision-makers, developing effective prototypes can be difficult due to a lack of tools, standards, and application-specific data.
\end{mybox}
\subsection{Evaluation}
\label{sec:evaluation}
\textbf{Evaluation} of FL systems is critical to convince clients to join the network by showing that an FL system will have desirable properties with respect to concerns such as accuracy, privacy, and intellectual property. However, not only are there multiple client organizations in cross-silo FL, but they may also have varying views on these concerns. Another important aspect of evaluation is communicating technical results to non-technical stakeholders, as evaluation requires stakeholders both quantify and address clients' concerns.

\textbf{Misaligned expectations lead to general challenges.} While clients are generally receptive to the idea that FL can improve outcomes over local training, a proof-of-concept is needed to give decision-makers a sense of {how} much value FL can bring. For decision-makers, a key challenge at this stage is clearly defining expectations and outcomes of FL, which requires a degree of technical expertise and clear communication with implementers. In addition to a basic understanding of the benefits of FL, stakeholders must also have foundational understanding of data-driven practices and ML systems. Unfortunately, in \ref{quote:data-maturity}, \pid{16} discussed that many organizations are simply not ready to pursue a project in ML, let alone FL.

\refstepcounter{quotecounter}\label{quote:data-maturity}
\begin{quotebox}
    {...before you can set up a federated network, you need to have your data ready, think about FAIR data principles, and so on. The reality is that often, in many companies still, data lives in Excel sheets.}{\pid{16}, on organization maturity needed for FL}
\end{quotebox}

Beyond improving over a local-only model, clients may not have a good sense of what to expect from FL. Unlike traditional ML where public datasets and established baselines provide clear reference points, FL performance depends on additional factors such as data heterogeneity and the aggregation algorithms employed. In \ref{quote:eval-centralized}, \pid{7} mentioned that client stakeholders wanted to see the performance of FL match that of centralized training. However, matching centralized training is often not possible from both a technical perspective (even in research settings) and a legal perspective (which prevents clients from training a central model to compare to).

\refstepcounter{quotecounter}\label{quote:eval-centralized}
\begin{quotebox}
    {A common thing is comparing to centralized...It's actually slightly annoying, even if they can't actually do this with centralized, there's often a feeling inside the organization that they want to see this be as good as centralized.}
    {\pid{7}, on clients' expectations of FL relative to centralized ML}
\end{quotebox}

Another aspect is tension between decision-makers and implementers, which stems from a gap in technical knowledge and interests. Decision-makers often lack a deep technical background in FL or ML and may struggle to grasp the nuances of various FL approaches. For example, others discussed how it can be difficult to communicate concepts to non-technical users (\pid{5}, \pid{11}). These may even be general to ML and not FL-specific concepts, such as recall and F1-score. Therefore, implementers face the challenge of translating complex technical concepts into business-relevant metrics and communicating the practical implications to non-technical stakeholders. In addition to bridging this knowledge gap, implementers often need to also perform extensive experiments to provide decision-makers with multiple options~(\ref{quote:eval-dp}).

\refstepcounter{quotecounter}\label{quote:eval-dp}
\begin{quotebox}
    {I think the main thing that would give us confidence is just testing different scenarios, testing different privacy settings, the different values of epsilon, different clipping values.}
    {\pid{17}, on the need for extensive testing}
\end{quotebox}

On the other hand, decision-makers must also play an active role in defining the evaluation framework. In \ref{quote:eval-communication}, \pid{20} discussed how a streamlined governance process and the ability of decision-makers/regulators to balance the data protection agenda with development is critical for broader adoption of FL.

\refstepcounter{quotecounter}\label{quote:eval-communication}
\begin{quotebox}
    {There's some examples of a regulator leading the way...the data fuels various development, so a regulator [has to have the] ability to balance the agenda between data protection and development, and then [also be] very technologically educated.}
    {\pid{20}, on privacy regulators and implementers}
\end{quotebox}


\textbf{Current networks rely on honest clients and legal contracts.} All subjects reported that their cross-silo FL projects operate within established organizational relationships governed by legal contracts. These contracts impose various agreements on details such as training protocol, data expectations, model access and usage, and penalties for violation. As a result, present real-world systems have a fundamentally different incentive structure compared to those studied in academic literature. 
While contracts do not provide any additional defense against potential attacks, they provide clients with a legal means of retaliation and sufficient level of trust to move forward with the collaboration. 

We observed that all cross-silo FL deployments have a single trusted entity handle the coordination of FL. The coordinator handles logistical tasks such as onboarding and contracting, as well as technical tasks such as model aggregation and monitoring. While this role is mainly carried out by a client who leads the consortium, platforms can also share same of these responsibilities. \pid{8} noted that while the star topology of this network is convenient for handling these concerns of coordination, it presents unique challenges for developing secure computation methods~(\ref{quote:trust-star}). More generally, this centralized approach raises questions about long-term sustainability and scalability, as the coordinator may become a single point of failure or a bottleneck for network growth.

\refstepcounter{quotecounter}\label{quote:trust-star}
\begin{quotebox}
    {I've noticed that the use cases are a bit harder if you involve external clients. For example, they really didn't want to have to talk to other clients. So they have to talk only with us.}
    {\pid{8}, on client preferences of network structure}
\end{quotebox}

In a larger consortium, a smaller, more experienced ``inner circle'' of organizations and individuals have a decision-making role~(\ref{quote:trust-circle}). The inner circle typically consists of organizations with the strongest technical capabilities or most significant investments in the collaboration.

\refstepcounter{quotecounter}\label{quote:trust-circle}
\begin{quotebox}
    {We call it inner circle and outer circle; so inner circle is sort of like the people who play more of a leadership role. Maybe they have more maturity...and therefore [have] strong suggestions...and [the outer circle] are just happy to participate because they see the benefits.}
    {\pid{20}, on decision-making in larger networks}
\end{quotebox}

Clients have varying views towards privacy, which leads to this observed dependence on a central coordinator. Discrepancies occur on many levels, ranging from the organization's location to its individual members. Several subjects mentioned that collaborations across states (provinces) in countries like the U.S. can be difficult due to varying state-level laws (\pid{1}, \pid{20}). This challenge also extends to international collaborations (\pid{17}), and handling regulations for each different industry (\pid{9}). 

In addition to regulatory differences across organizations, the individuals at these organizations themselves may have different views towards privacy. Some people are very attracted to FL's potential to overcome regulatory limitations, while others may care highly about the intentions of these regulations and are thus resistant to such efforts (\pid{7}). Since each legal team needs to give their approval, disagreements on the intra-organization level blocks progress on the inter-organization level (\pid{16}). Even in non-FL collaborations between a client, platform, and service vendor, clients have varying preferences for where to keep their data and perform the service: entirely local, on the mutually trusted platform, or directly on the vendor's servers (\pid{19}). Finally, FL must abide by multiple data privacy regulations especially when the clients exist internationally, such as GDPR and EU AI, adding additional complexity over data-driven solutions which do not rely on model training (\pid{4}).

Due to these factors, subjects noted that platforms and clients take a practical approach towards privacy and trust, emphasizing aspects of transparency, accreditation, and governance (\ref{quote:eval-system}). Therefore, while some subjects discussed ongoing work in privacy-enhancing technologies like homomorphic encryption and differential privacy, others mentioned that these techniques were not necessary for their use case.

\refstepcounter{quotecounter}
\label{quote:eval-system}
\begin{quotebox}
    {We have different [pieces] that can be used to convince people...just the clarification that in the whole use case, there's no data leaving...then we have a bunch of accreditations, we have a governance and regulatory affairs person...on the technical level, we have a few measures...In that sense of, privacy-enhancing technologies, to give comfort...we're not really using those at the moment, because the nature of the federation and the collaborations that we're running don't require it.}
    {\pid{21}}
\end{quotebox}


\textbf{Incentives restrict network diversity.} Intellectual property is a major concern in industries where the client's data provides them with a competitive advantage. Therefore, while FL can improve model performance, it has business risks which are hard to quantify and may outweigh these benefits. One of the major concerns clients reported was that clients should equally (or fairly) benefit from the final model. Even if FL improves model performance for all clients, their individual benefits are often inversely proportional to their contribution: the clients with the least or lowest-quality data stand to gain the most from a shared model. Furthermore, this notion of ``equal benefits'' depends on how fairness and benefits are quantified. As a result, FL collaborations tend to only naturally form across organizations which are of a similar size, where fair outcomes are both easier to define and achieve. 

While algorithms for contribution estimation and incentive mechanisms exist \cite{tang2021incentive,karimireddy2022mechanisms,huang2023promoting,chen2024free}, clients and platforms do not actively consider these methods. As mentioned above, FL deployments are currently established in a way where clients are of roughly equal size. When clients have specific concerns, these typically do not fall nicely under existing incentive frameworks and need to be negotiated on a case-by-case basis (\ref{quote:incentives-valuation}). 
Platforms largely do not provide ways to address such issues; in the one case where contribution estimation was used in a real-world setting by \pid{10}, a group of academic researchers estimated contributions using a system that was decoupled from the clients' network.

\refstepcounter{quotecounter}\label{quote:incentives-valuation}
\begin{quotebox}
    {[Data valuation] probably needs to be figured out case by case, and it's probably best to approach it from 'what is the benefit or potential benefit to all members?' first, and then, 'do we we need to balance it in such a granular fashion?'}{\pid{16}, on the potential benefit of data valuation}
\end{quotebox}

Furthermore, as discussed by \pid{21} in \ref{quote:incentives-outcomes}, FL requires ``reverse-engineering'' the exact outcome that clients want to achieve. However, different clients often have competing objectives, making it challenging to satisfy all clients simultaneously. Additionally, clients face uncertainty about outcomes due to aforementioned evaluation challenges: translating system metrics to anticipated business impact is difficult.

\refstepcounter{quotecounter}\label{quote:incentives-outcomes}
\begin{quotebox}
    {What federation forces you to do is to think, very clearly, [and] up front, about what's exactly going to happen, right? And then you kind of, in a sense, reverse engineer the technical stuff to make sure that a particular outcome can happen.}{\pid{21}, on clients' expectations of FL}
\end{quotebox}

\begin{pbox}{Open Problem}
\label{problem:standards_evaluation}
    {To accelerate network formation, clients need guidelines on how to evaluate FL systems. Furthermore, it is important to understand how different relationships between clients affect feasibility and long-term growth.}
\end{pbox}


\textbf{Platforms can assist with both technical and human aspects of evaluation.} By offering accessible tools and standardized protocols, platforms can significantly reduce the complexity of FL evaluation for individual organizations. First, platforms can provide technical assistance through FL simulation and sandboxing tools (\pid{15}, \pid{20}). These tools enable clients to test different FL approaches in controlled environments before full deployment, allowing them to evaluate system components without requiring full coordination with other clients. Subjects also emphasized the importance of interface-only, no-code solutions for non-technical users. Even for complex issues such as fairness and incentives, \pid{4} hypothesized that a convenient interface could be more effective than a sophisticated algorithm at convincing decision-makers of the value of FL (\ref{quote:eval-interface}).

\refstepcounter{quotecounter}
\label{quote:eval-interface}
\begin{quotebox}
    {I think in practice, what we've seen at least a lot of times, what matters is the way it's shown to the real user. I mean, you may have a very nice algorithm to guarantee some good participation, or is equal in a certain way, but I think it ultimately comes down to how the user would bid ... that kind of a platform, I don't think it exists.}{\pid{4}, {on how human preferences are underexplored in FL.}}
\end{quotebox}

\begin{pbox}{Open Problem}
\label{problem:incentives_human}
    {Due to human factors affecting clients' willingness to participate in FL, future research should consider how humans interact with FL systems, e.g., presenting results in a way that incentivizes stakeholders to join.
    }
\end{pbox}

Furthermore, long-lasting networks can help establish trust. Once a pilot study is successful, it shows that trustworthy FL collaboration is possible, and it provides a standing investment so new clients do not need to repeat the difficult process of establishing evaluation standards~(\ref{quote:trust-scale}).
\refstepcounter{quotecounter}\label{quote:trust-scale}
\begin{quotebox}
    {That is the hope, just for kind of more, streamlined experience and processes. Because, it doesn't make sense for, like, newcomers to come and then kind of go through all the things that we have gone through, right?}
    {\pid{20}, on how network-wide evaluation standards ensure a streamlined experience for new clients.}
\end{quotebox}

Lastly, clients may be unwilling to place their trust in any one entity and are looking forward to future cryptographic technologies that can enable completely untrusted parties to collaborate (\ref{quote:trust-blockchain}). In addition to the limitations around honest network formation, these technologies often come with significant computational overhead and complexity that limits their adoption in current systems~\cite{zhu2023blockchain,xie2024efficiency,xu2025secure}.

\refstepcounter{quotecounter}\label{quote:trust-blockchain}
\begin{quotebox}
    {Eventually, I don’t think people would like to trust a single party. Maybe, in some cases, there is a large party who would behave honestly. But in examples where people may not want to have a central party, that's where the blockchain can play a role.}
    {\pid{13}, on the structure of future networks}
\end{quotebox}


\begin{mybox}{Takeaway}
FL network formation relies on \textbf{evaluating} multiple aspects beyond model accuracy. Unfortunately, a lack of resources and relevant standards makes it difficult for decision-makers to resolve such issues.
\end{mybox}
\subsection{Deployment}
\label{sec:deployment}
Once the three prior stages (ideation, prototyping, evaluation) are complete, all clients have agreed on the design of the FL system. At this point, the system must be \textbf{deployed} to the remote client sites to train the final model. Subjects emphasized key factors to streamline deployment, such as wide compatibility support and tools to convert code from simulation to deployment. These challenges included both setting up infrastructure and normal system operation, though subjects tended to focus more on the former.

\textbf{FL deployment requires diverse expertise.} Almost all subjects mentioned that a diverse set of knowledge is needed to handle FL systems. While ML and FL engineering skills are certainly needed for FL projects, subjects also noted that projects rely just as much on non-ML knowledge, such as in computer systems and networks. Furthermore, few individuals have such diverse expertise or the bandwidth to work on multiple components (\ref{quote:deploy-expertise}).

\refstepcounter{quotecounter}
\label{quote:deploy-expertise}
\begin{quotebox}
    {In a production setup after doing the development work, a lot of the time, these researchers and engineers are not fit to do that type of infrastructure.}{\pid{15}, on the expertise needed for deployment}
\end{quotebox}

\textbf{Software should be designed with deployment in mind.} Similar to the ideas around prototyping (Section~\ref{sec:prototyping}), compatible and easy-to-use tools are important for deployment. Subjects noted that debugging during the deployment stage can be particularly difficult, as developers usually have restricted access to  an external client's system. This challenge is most extreme in networks which do not have the assistance of a platform; here, not only are there issues with debugging across systems, but also w.r.t. confidence in external results even if no error was encountered~(\pid{1}, \pid{2}). 

Platforms play a key role in providing production-ready software and support. However, since platforms are unable to cover all types of specifications in a single product, 
it is also important to address these concerns on an individual project basis when possible. For example, as discussed by \pid{12} in \ref{quote:deploy-access}, system designers can implement automated configuration mechanisms such as network discovery. While setting up these features has its own additional set of challenges, these mechanisms can reduce or remove the need for additional reconfiguration as the system is updated.

\refstepcounter{quotecounter}
\label{quote:deploy-access}
\begin{quotebox}
    {We have developed a self discovery mechanism with broadcasted packages, etc, in order to establish [the network] as simply as possible, and make it easy for the end user. But [the hospital IT departments] are not willing at all to provide admin access for some of them. Probably this is still an ongoing process ... I think that some fellow researchers will have to go on site, at least for some of the hospitals.}{\pid{12}, on network security at hospitals.}
\end{quotebox}

\textbf{Systems require maintenance after deployment.} Once infrastructure is set up, there are additional questions around normal operation of the system. The most important of these is usage and ownership of the model. As discussed in Section~\ref{sec:evaluation}, current networks are typically formed by clients who expect roughly equal benefits from FL. Unfortunately, networks take a simple but restrictive solution by requiring all clients to make roughly equal contributions and then sharing the final model with all clients. While several subjects expressed interest around more flexible frameworks of model ownership, these were discussed in other collaborative learning contexts, such as secure inference and distributed training.

Another key issue is that clients need to update the system over time. This can be for a variety of purposes, such as to train on new data, fix errors, or satisfy downstream specifications. While organizations may be experienced with maintaining traditional ML systems, the distributed nature of FL introduces new challenges. For example, systems which are continuously updated on new data need specific approval. When discussing how models from FL would be used in production, \pid{13} explained that companies in regulated industries have existing systems around model governance, and that human aspects such as approval and monitoring were the main difficulty~(\ref{quote:deploy-monitor}). In the context of FL, this burden of approval is multiplied across multiple organizations, who may not be comfortable with the ability of external parties to modify their training data. Therefore, FL systems must not only adopt tools from MLOps to address traditional challenges of model deployment, but also build on these tools to address FL-specific concerns around trust and safety.

\refstepcounter{quotecounter}
\label{quote:deploy-monitor}
\begin{quotebox}
    {The system is in place for updating models, but it has to be approved. Even if it's approved, you have to continuously monitor it in case it meets a certain metric in testing but then doesn't meet that after approval.}
    {\pid{13}, on governance of deployed ML models}
\end{quotebox}

\begin{pbox}{Open Problem}
\label{problem:systems_monitoring}
{To address heightened concerns around trust and safety, cross-silo FL should adopt and build on tools from MLOps for monitoring, testing, and versioning.}
\end{pbox}
Lastly, subjects had various concerns about computation and communication costs. On the platform side, there is a challenge of monitoring usage and assigning costs correctly to users performing remote computation (\pid{5}). As for clients, there tended to be few concerns with the scale of compute. As discussed by \pid{17}, this is because applications were generally small-scale in most aspects, such as in the model, local dataset, or number of clients (\ref{quote:deploy-compute}). A few subjects (\pid{9}, \pid{18}) mentioned efforts to develop homomorphic encryption methods, but focused on how they are too expensive to even consider for model training.

\refstepcounter{quotecounter}
\label{quote:deploy-compute}
\begin{quotebox}
    {With [our domain] being relatively small data sets like dozens of maybe low hundreds per institution, then the data is not so big that the computational overhead will be, you know, too onerous.}{\pid{17}, on small computational costs}
\end{quotebox}

\begin{mybox}{Takeaway}
While platforms already solve key challenges with \textbf{deployment}, the structure of current FL networks restricts the types of challenges that arise in practice. Therefore, practitioners need to develop guidelines on how to maintain end-to-end FL systems.
\end{mybox}
\section{Discussion}
\label{sec:discussion}
\subsection{Insights On The Gap Between Research and Practice}
We now discuss potential reasons for the gap between research and practice of cross-silo FL. Based on our analysis, the following primary factors contribute to this misalignment.

\textbf{Cross-silo FL applications are highly diverse.} The cross-silo FL setting exhibits heterogeneity across multiple dimensions due to diverse and numerous types of client industries (\ref{quote:applications}). Unfortunately, the challenges and necessary solutions also differ across applications, making it difficult to establish a clear set of research priorities for the field. Even though the current applications of cross-silo FL are predominantly in healthcare and finance, the specific areas these applications explore are quite varied (e.g. genomics, radiology, and lab tests)~\cite{heyndrickx2023melloddy,soltan2024scalable,asti2024artificial}. 
This is in contrast with cross-device FL, where technology companies often focus on applications deployed in a mobile device environment, making it easier to integrate cross-device FL and establish clear research directions~\cite{yang2018applied,bonawitz2019towards,paulik2021federated,stojkovic2022applied,huba2022papaya,ji2025private}. 

\refstepcounter{quotecounter}
\label{quote:applications}
\begin{quotebox}
    {So. as you you probably know, federated learning actually started on edge devices, right? Mobile phones like, that's an original use case. But since then, I think a lot of the real world applications in academia and also industry, have been on the so-called cross silo use case where you have different hospitals or financial institutes working together.}{\pid{9}, on real-world FL applications}
\end{quotebox}

\textbf{Cross-silo applications have significantly less investment.} Cross-silo FL is naturally more applicable to small and medium-sized client (organizations), as they stand to benefit more from external collaboration. Unfortunately, these clients have significantly less research investment in cross-silo FL, which leads to poorly defined research agendas, a lack of research in and advertisement of practical challenges, and a lack of public infrastructure. Echoing this point in \ref{quote:resources}, \pid{11} believed that the first major cross-silo FL application in finance would have to come from two large banks.

\refstepcounter{quotecounter}
\label{quote:resources}
\begin{quotebox}
    {I also don't think that a small company will be the one who takes this first step because they have no data, and they have to talk to all the stakeholders, and they have no influence, no big standing. So I think it will be maybe two big banks, that just decide, well, let's try this out.}{\pid{11}, on how limited resources prevent small organizations from pursuing cross-silo FL.}
\end{quotebox}

This is in contrast to cross-device FL, where there is large-scale research investment by technology giants and the research agendas and practical challenges are well-articulated and advertised~\cite{kairouz2019advances,bonawitz2019towards,paulik2021federated}.

\textbf{Literature conflates Cross-\textit{Silo} FL with Cross-\textit{Device} FL.} 
In~the FL literature, ``cross-silo FL'' is sometimes used in the same way as ``cross-device FL'' without fully clarifying the distinction between the two. First, we discuss this trend in FL survey papers. While there are some survey papers on cross-silo FL which clearly outline relevant challenges~\cite{huang2022cross,huang2023promoting,korneev2025survey}, most survey papers in FL place significantly less emphasis on cross-silo FL. General FL survey papers do identify cross-silo as an important area of FL, but do not consider in detail the specific challenges that the structure of cross-silo FL brings with it~\cite{kairouz2019advances,rahman2023federated,joshi2022federated,vzalik2023review,liu2024vertical}. Other survey papers discuss FL without referring to either cross-device or cross-silo; in these papers, the contents can hint more towards either setting~\cite{wen2023survey,teo2024federated,eden2025scoping,shi2023responsible,guan2024federated,woisetschlager2024survey,pei2024review}. 
Second, we observe that papers may include ``cross-silo'' in their title or abstract while focusing on challenges that are broadly applicable to FL in general, such as communication~\cite{majeed2020cross,marfoq2020throughput,chen2020dealing,huang2021personalized,lin2021semifed,fu2021vf2boost,luo2022adapt,liu2022fedbcd,xu2022closing,xu2022coordinating,wu2023faster,qin2023fedapen,aggarwal2023federated,qi2023cross}. Furthermore, many papers do not specify either of the terms. While such papers tend to be more applicable to cross-device FL than cross-silo FL,  this lack of specificity makes it difficult to understand which setting these methods are designed for. 

Combined with the findings from our interviews, this raises an interesting disconnect: the two settings are treated similarly in research, but their challenges are significantly different in practice. In order for research in cross-silo FL to be practical, researchers must clearly define how its challenges differ from those of cross-device settings; one example is the user-level vs. example-level definition of differential privacy~\cite{mcmahan2018learning,liu2022privacy}. 

We posit that the preference for cross-device as the ``default'' FL setting stems from the two factors (of application heterogeneity and investment) discussed earlier. Additionally, \pid{7} mentioned that the similarity between the terms ``cross-device" and ``cross-silo" may have also contributed to this trend.
Moreover, in \ref{quote:naming}, \pid{7} felt that the two settings could "have separate names quite comfortably" given that "the challenges are so different". This observation highlights a humorous possibility: that a significant portion of the research-practice gap in cross-silo FL might be ameliorated simply through clearer terminology.

\refstepcounter{quotecounter}
\label{quote:naming}
\begin{quotebox}
    {I think people feel, outside of [technology companies], people who have done things, I think everyone else is much more comfortable with cross-silo...We could have separate names quite comfortably for the two things. Obviously, they're very related. But the challenges are so different.}{\pid{7}, comparing cross-device vs. cross-silo FL}
\end{quotebox}

\subsection{Solutions and Open Problems}

\textbf{Settings where cross-silo FL is more likely to be realized.} There is a need to think more carefully about the settings where cross-silo FL is applicable and where issues of incentives are less severe. As a few example settings, \pid{10} mentioned that it could be in the interest of two small competitors to collaborate against a third larger competitor. \pid{11} noted that there are less competing incentives in vertical and multi-modal FL, where organizations possess data from different domains and thus are more likely to have non-competing business objectives~\cite{liu2024vertical,khan2025vertical}. Similarly, \pid{13} noted that incentives would be more aligned in social good applications as opposed to monetary applications. 

\textbf{Expanding the scope of FL.} FL practitioners should be thinking about broader problems, particularly beyond model training (\ref{quote:scope}). Subjects were curious about how to use FL with other ML workflows, such as RAG (\pid{2}) and automated data labeling (\pid{3}). Furthermore, stakeholders noted that FL in the context of training models is difficult, but simpler use cases may be attainable. For example, \pid{2} worked on a vision application, but believed it would still be interesting and even more valuable to focus on easier tabular problems with fewer features. Similarly, \pid{1} and \pid{17} worked on projects using federated analytics, and mentioned that evaluation challenges would have been much more difficult if they had considered (multiple rounds of) model training. Platforms (\pid{5}, \pid{7}, \pid{9}, \pid{15}, \pid{16}) were also aware of this demand for tools beyond model training and therefore emphasized the importance of a general and flexible design~\cite{roth2022nvidia,eichner2024confidential,stricker2024fl}.

\refstepcounter{quotecounter}
\label{quote:scope}
\begin{quotebox}
    {I feel the scope of federated learning probably could be made a bit bigger, and look into some new approaches. Because, if you think about it, nothing's model training these days.}{\pid{4}, on expanding the scope of FL}
\end{quotebox}

\textbf{Opt-out guarantees.} Several researchers (\pid{3}, \pid{6}, \pid{18}) noted the importance of opt-out guarantees, where the final model can be updated in a modular way if a client decides they no longer wish to participate in the network. For example, one-shot model merging methods can be adapted to perform FL in a single round of aggregation~\cite{ilharcoediting,shi2025flexolmo,kuo2025exact}. Unlike standard multi-round FL training, this avoids the need to restart FL from scratch, as the opt-out client's model parameters can easily be unmerged (e.g. subtracted) from the aggregate (\ref{quote:modular}). 

\refstepcounter{quotecounter}
\label{quote:modular}
\begin{quotebox}
    {When pretraining a language model, it's hard to make a one-time decision on what data to include or not. Allowing you to use data in a more modular fashion is useful and enables more organizations to participate.}{\pid{3}, on the benefit of opt-out guarantees}
\end{quotebox}

\textbf{Data marketplaces.} While data marketplaces for FL have not been realized in practice, more general marketplaces for ML  are taking shape~\cite{amazonHealthcareDatasets,healthverityHealthVerityMarketplace,prognoshealthHome,marcheMarcheHealth,finazonFinazonMarketplace}. These services connect data owners to service vendors, which reduces the cost of running pilot studies and engineering on the data owner's end. In several interviews, subjects (\pid{4}, \pid{9}, \pid{10}, \pid{20}) expressed interest in a data marketplace connecting multiple data owners. Such a marketplace would have several benefits: it would establish standards around data privacy and valuation, give data owners flexible control over their prices, and ultimately unlock more cross-organization collaborations.

\textbf{FL needs broader standards.} In practice, standards are a critical bottleneck across all stages of cross-silo FL projects. Unlike cross-device FL, where a single organization can establish its own internal protocol, cross-silo collaborations require agreement between multiple independent entities with varying capabilities, regulations, and incentives. The absence of standardized approaches forces each collaboration to negotiate standards from scratch, such as system design, privacy requirements, and client compensation. This lack of standardization significantly increases the time and cost of establishing cross-silo FL networks, as each new collaboration must reinvent solutions to common problems. 

Developing cross-silo FL standards requires coordination between multiple stakeholders, including industry particpants, technology platforms, and regulatory bodies. While these coordination challenges are difficult to overcome, there are promising examples of regulators taking a leadership role in coordinating between industry, government, and public consultation processes (\ref{quote:regulator}). However, in order to handle heterogeneous cross-silo applications, standards must balance specificity with flexibility. Without such standards, adoption of cross-silo FL will remain limited to major companies, limiting the field's potential impact.

\refstepcounter{quotecounter}\label{quote:regulator}
\begin{quotebox}
    {The privacy law in [Country] is very fragmented...there's all this variation, and then privacy being the provincial jurisdiction, and then they have to work with the federal counterparts. So coordination is very difficult in that regard. So I think there's a number of issues in terms of, you know, why certain emerging technology adoption can take so long, because in [Country], it just takes so much consultation and coordination to get to that state.}
    {\pid{20}, on how coordination over privacy standards slows down cross-silo FL adoption}
\end{quotebox}

\textbf{Improving awareness and communication.} Subjects agreed general awareness about FL is an issue, but also expressed hope that, with the right approach, they can successfully communicate concepts of FL and other related fields. In \ref{quote:awareness}, \pid{5} discussed how people facilitating platform-client collaborations have both business and technical expertise. 

\refstepcounter{quotecounter}\label{quote:awareness}
\begin{quotebox}
    {It's often the people whose data is going to be used that have no idea what federated learning is...our sales team is not your typical sales team, you know. It's people with PhDs and Master's degrees that are that are doing selling because they really have to be able to explain really complicated concepts to people in a cogent way.}{\pid{5}, on customer knowledge of FL}
\end{quotebox}
\section{Conclusion}
In this paper, we presented results from semi-structured interviews with 21 cross-silo FL stakeholders to explore the disconnect between research and practice of cross-silo FL. We found that there is a significant misalignment between the challenges studied in these two settings; while research focuses heavily on technical challenges such as communication efficiency and data heterogeneity, practitioners struggle more with organizational barriers such as lack of FL awareness, establishing trust between competitors, and navigating complex legal requirements. To bridge this gap, we identify several promising research directions including developing industry-wide standards, creating accessible tools for non-technical stakeholders, and expanding FL research beyond model training to address the full deployment pipeline. Our findings suggest that for cross-silo FL to achieve its promise, the research community should realign its priorities to better address the practical challenges that prevent real-world adoption.
\\
\\
\\
\\
\textbf{Limitations.} We note that we consider a mostly US-centric perspective, as shown in Table~\ref{tab:subjects}. Additionally, we mostly report on applications in healthcare and finance, as we found these to be the most dominant applications. Our interviews consider general challenges in cross-silo FL collaborations, rather than technical challenges unique to implementers and researchers. Therefore, we interview a diverse set of subjects with varying degrees of technical experience in FL. 

\pagebreak



\bibliography{main}
\clearpage
\pagebreak

\appendix
\phantomsection
\section*{Appendices}
\addcontentsline{toc}{section}{Appendices}
\let\oldsection\section
\renewcommand{\section}[1]{%
  \refstepcounter{section}%
  \addcontentsline{toc}{subsection}{\thesection\quad #1}%
  \oldsection*{\thesection\quad #1}%
}
\renewcommand{\thesection}{\Alph{section}}

\section{Interview Questions}
We list the questions we used to guide our interviews. However, we note that the interviews were semi-structured and the interview questions were varied dynamically based on the conversation and the specific roles of the subjects.

\textbf{List of interview questions (for clients).} 
\label{appendix:questions}
\begin{enumerate}
    \item \textbf{What ML problem (or problems) are you trying to solve?}
    \begin{enumerate}
        \item What is the modality of your data? (image, text, audio, etc.)
        \item What is the modeling objective? (classification, regression, etc.)
        \item What types of models are you using? (linear models, decision trees, neural nets)
    \end{enumerate}
    
    \item \textbf{How is your local model development pipeline set up?}
    \begin{enumerate}
        \item What libraries do you use to train models?
        \item What format do you use to store models?
        \item What computational resources does your organization allocate for modeling?
    \end{enumerate}
    
    \item \textbf{How would cross-silo FL improve model development?}
    \begin{enumerate}
        \item What is the nature of individual silos? (e.g. external organizations, internal sites)
        \item What restrictions are there on sharing data between silos?
        \item What incentives do individual data silos have to collaborate? (or what incentives can you provide?)
        \begin{enumerate}
            \item Is the data/computation from a single silo (in)sufficient?
            \item Is data at a particular silo more valuable?
            \item Are there issues of unfair / skewed performance across silos?
            \item Are certain silos at more risk of privacy attacks?
            \item Are you concerned about intellectual property and not being recognized for your contribution?
            \item How do you establish trust between silos?
        \end{enumerate}
    \end{enumerate}
    
    \item \textbf{What challenges do you face in implementing cross-silo FL?}
    \begin{enumerate}
        \item How is training coordinated across multiple silos?
        \begin{enumerate}
            \item Is training synchronous, asynchronous, or something else?
            \item How much communication is done (e.g. number of aggregation rounds)
        \end{enumerate}
        \item What system differences across silos might limit FL feasibility?
        \begin{enumerate}
            \item How are models (or data) shared? In what format?
        \end{enumerate}
        \item How heterogeneous is the data in different silos? (labels, features, etc?)
        \item How does the training procedure differ across silos?
        \item Expectations around privacy
        \begin{enumerate}
            \item What risks are you aware of / are communicated to you?
            \item Is FL alone sufficient to protect privacy, or are other techniques needed?
            \item Are users ready to trade privacy for utility and efficiency?
            \item How much of it is trust in partners vs legal contracts?
        \end{enumerate}
    \end{enumerate}
    
    \item \textbf{How do you maintain models after development and/or deployment?}
    \begin{enumerate}
        \item Who owns the final model(s)?
        \item Is a single model applied at all silos? Or a personalized model for each silo?
        \item Who is able to use the models, and what type of access do they have?
        \item When do you re-use or build upon internal models?
        \item If a single silo updates their data, how should the other silos respond? In general, when do you determine when to train a new model and repeat the entire cross-silo FL procedure?
    \end{enumerate}
\end{enumerate}

\clearpage
\textbf{List of interview questions (for platforms).} 
\begin{enumerate}
    \item \textbf{Describe your organization and role.}
    \begin{enumerate}
        \item What is your overall mission, and how relevant is it to cross-silo FL?
        \item What is the scale of your organization / product (e.g. number of clients, developers, managers)
        \item Does your software offer features unique to cross-silo applications?
    \end{enumerate}
    
    \item \textbf{What are some of your largest clients or most common applications?}
    \begin{enumerate}
        \item Do your clients focus on a particular domain? (i.e. data, task, model)
        \item How does your broad or specific focus affect your software design?
        \item Participation from underdeveloping nations
    \end{enumerate}
    
    \item \textbf{Which features are most important?}
    \begin{enumerate}
        \item ML framework support
        \item guardrails in practice
        \item efficient training
        \item communication across distinct nodes (for real-world applications)
        \item local simulation (for research purposes)
        \item FL algorithm implementation (e.g. FedProx, FedOpt)
        \item domain-specific analysis tools
    \end{enumerate}
    
    \item \textbf{How do you decide which features to support?}
    \begin{enumerate}
        \item online / public requests (e.g. Github issues)
        \item talking directly with clients
        \item observing ongoing research
    \end{enumerate}
    
    \item \textbf{What makes users choose your service?}
    \begin{enumerate}
        \item Obvious benefits: enable organizations to use a common framework, provides boilerplate code
        \item Open-source software / documentation
        \item How familiar are your clients with these technologies (general software / programming knowledge, ML, FL, distributed systems)
    \end{enumerate}
    
    \item \textbf{What challenges are there in adoption?}
    \begin{enumerate}
        \item Are there gaps in what clients think FL will solve versus what it actually delivers?
        \item How do you demonstrate the impact / benefit of FL?
        \item Do you encounter additional issues of heterogeneity? e.g. different organizations want to use different model architecture.
        \item Why may companies be reluctant to pursue FL?
        \item How do legal and contractual aspects handle these concerns?
    \end{enumerate}
    \item \textbf{Any guiding principles / design patterns that are unique to FL software?}
\end{enumerate}

\clearpage
\begin{figure}
    \centering
    \includegraphics[width=8.8cm]{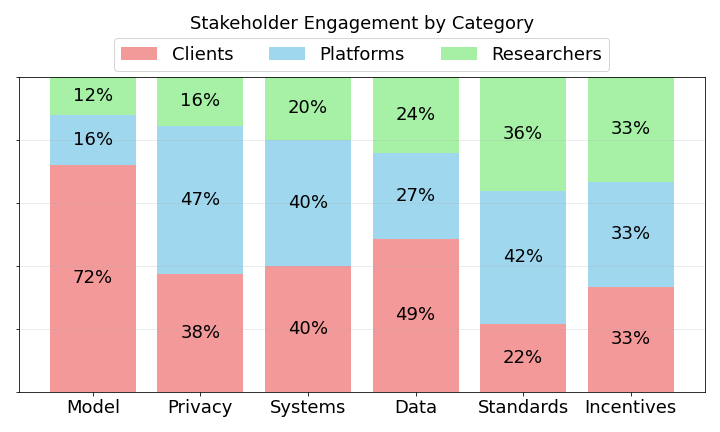}
    \caption{\textbf{We compare the number of stakeholders who mentioned each of our 6 main topics during the interview.} We find that each stakeholder group emphasizes different topics (relative to other stakeholders): clients care about models and data while platforms care about systems, privacy, and standards. Percentages are normalized by both the stakeholder group size and total times the topic is mentioned i.e. equal percentages mean an equal fraction of each stakeholder group, not an equal number of stakeholders.}
    \label{fig:topics_stakeholders}
\end{figure}

\begin{table*}[h]

\centering
\begin{tabular}{lll}
\toprule
\textbf{Category} & \textbf{Subcategory} & \textbf{Description} \\
\midrule
Data & heterogeneity & Heterogeneous data distributions across clients harms FL optimization \\
Data & harmonization & Data pipeline challenges and discrepancies across organizations \\
General & platform & Scoping the role of platforms for coordination and infrastructure support \\
General & awareness & Improving awareness and understanding of FL in the community \\
Privacy & legal & Working with regulators; gaps between research and regulations in practice \\
Privacy & leakage & Risks of information leakage from models \\
Model & communication & Communication efficiency \\
Model & aggregation & Aggregation algorithms \\
Incentives & valuation & Valuation of contributions, fair compensation \\
Incentives & competition & Negative concerns with other participants \\
Systems & compute & Computational resources \\
Systems & infrastructure & Setting up software and compute infrastructure \\
\bottomrule
\end{tabular}
\caption{Selected challenges from literature and practice}\label{tab:categories}
\end{table*}

\section{Identifying Challenges Discussed in Interviews}
Once the interviews were completed, researchers reviewed the transcripts to identify parts which are (1) emphasized by subjects or (2) relevant to ongoing research in cross-silo FL. To do this, researchers manually coded (labeled) the transcripts. From an initial pass over the transcripts, we identified 6 key categories of discussion: on modeling, privacy, systems, data, standards, and incentives. In Figure~\ref{fig:topics_stakeholders}, we show the relative frequency at which each type of stakeholder (client, platform, researcher) is focused on that category. On a second pass over the transcripts, we identified more fine-grained categories which correspond to particular challenges (e.g. model communication and model aggregation).

\section{Literature Analysis Technique Leading to Fig.\ref{fig:topics_abstracts}} 
\label{appendix:interview_plot}
We discuss our literature analysis technique leading to Figure \ref{fig:topics_abstracts}. To determine popular challenges in the literature, we scrape the top 200 cited papers on Semantic Scholar corresponding to the term ``cross silo federated learning''. We then iterated on a set of challenges and keywords for each challenge. The process of categorizing the abstracts involved reviewing abstracts by hand one-by-one and determining their relevant topic-keyword pairs. Additionally, each newly added keyword must be reviewed to confirm that it produces no false positives; otherwise, a more specific keyword is chosen instead. 

Next, to show differences on both a coarse and fine-grained scale, we identified 6 areas these challenges (from both abstracts and interviews) fall under: data, model, privacy, incentives, systems, and general. Within each area, we filter 2 challenges, the most frequent challenge from the abstracts and interviews respectively, for a total of 12 challenges.

\section{Additional Subject Details}
In 11 out of the \NN interviews, our subjects discussed successfully deployed systems, while the other subjects only discussed their experiences with systems prior to deployment. We indicate this as the \textbf{stage} column in the table. We also considered subjects from both the USA and international sites, which is shown by the \textbf{site} column.

\end{document}